\documentclass{llncs}
\usepackage{CJK}
\usepackage{graphicx}
\usepackage{bm}
\usepackage{lipsum}
\usepackage{url}
\usepackage[colorlinks=true,linkcolor=black,urlcolor=blue,backref]{hyperref}
\begin{document}
%
\title{DiaKG: an Annotated Diabetes Dataset for Medical Knowledge Graph Construction}
%
%
%
%
\author{
    Dejie Chang\inst{1}
    \and
    Mosha Chen\inst{2} 
    \and
    Chaozhen Liu\inst{1}
    \and
    Liping Liu\inst{1}
    \and
    Dongdong Li\inst{1}
     \and
    Wei Li\inst{1}
    \and
    Fei Kong\inst{1}
    \and
    Bangchang Liu\inst{1}
    \and
    Xiaobin Luo\inst{1}
    \and
    Ji Qi\inst{3}
    \and
    Qiao Jin\inst{3}
    \and
    Bin Xu\inst{3}
}
\authorrunning{F. Author et al.}
%
\institute{
  Miao Health  
\email{$\{$changdejie,liuchaozhen,liuliping,lidongdong,liwei,kongfei, liubangchang,luoxiaobin$\}$@miao.cn}
\and
 Alibaba Group 
\email{chenmosha.cms@alibaba-inc.com}
\and
Tsinghua university
\email{$\{$jqa14,qj20$\}$@mails.tsinghua.edu.cn,xubin@tsinghua.edu.cn}
}
\maketitle              
\begin{abstract}
	
Knowledge Graph has been proven effective in modeling structured information and conceptual knowledge, especially in the medical domain. However, the lack of high-quality annotated corpora remains a crucial problem for advancing the research and applications on this task. In order to accelerate the research for domain-specific knowledge graphs in the medical domain, we introduce DiaKG, a high-quality Chinese dataset for Diabetes knowledge graph, which contains 22,050 entities and 6,890 relations in total. We implement recent typical methods for Named Entity Recognition and Relation Extraction as a benchmark to evaluate the proposed dataset thoroughly. Empirical results show that the DiaKG is challenging for most existing methods and further analysis is conducted to discuss future research direction for improvements. We hope the release of this dataset can assist the construction of diabetes knowledge graphs and facilitate AI-based applications.

\end{abstract}
\section{Introduction}

Diabetes is a chronic metabolic disease characterized by high blood glucose level. Untreated or uncontrolled diabetes can cause a range of complications, including acute ones like diabetic ketoacidosis and chronic ones such as cardiovascular diseases and diabetic nephropathy. With the rapid economic developments and changes in lifestyle, China has become the country with the most diabetes patients in the world: the prevalence of diabetes in Chinese adults is about $11.2\%$ and still increasing\cite{BMJ}. The medical expenses from diabetes without complications already account for $8.5\%$ of national health expenditure in China\cite{J}. Cardiovascular diseases, one of the complications of diabetes, are the leading cause of death in China. Diabetic nephropathy, another diabetes complication, could “waste the wealth that we’ve accumulated over the past 30 years in the drains of dialysis machines” according to \cite{CDIA}. As a result, diabetes is a serious public health problem in the realization of “Healthy China 2030” that requires interdisciplinary innovations to solve.

Knowledge Graph (KG) has been proven effective in modeling structured information and conceptual knowledge, especially in the medical domain\cite{KG}. Medical knowledge graph is attracting attention from both academic and healthcare industries due to its power in intelligent healthcare applications, such as clinical decision support systems (CDSSs) for diagnosis and treatment\cite{KNEE,KGE}, self-diagnosis utilities to assist patient evaluating health conditions based on symptoms\cite{TANGH,GANN}. High-quality entity and relation corpus is crucial for constructing knowledge base, however, there is no dataset dedicated to the diabetes disease at the moment. To address this issue, we introduce DiaKG, a high-quality Chinese dataset for Diabetes knowledge graph construction. 

The contributions of this work are as follows:

1. To the best of our knowledge, this is the first diabetes dataset for medical knowledge graph construction at home and abroad. 

2. In addition to the medical experts, we also introduce AI experts to participate in the annotation process to provide data insight, which improves the usability of DiaKG and finally benefits the end-to-end model performance.

We hope the release of this corpus can help researchers develop knowledge bases for clinical diagnosis, drug recommendation, and auxiliary diagnostics to further explore the mysteries of diabetes. The datasets are publicly available at \url{https://tianchi.aliyun.com/dataset/dataDetail?dataId=88836}

\section{DiaKG Construction}

\subsection{Data Resource}
The dataset is derived from 41 diabetes guidelines and consensus, which are from authoritative Chinese journals covering the most extensive fields of research content and hotspot in recent years, including clinical research, drug usage, clinical cases, diagnosis and treatment methods, etc. Hence it is a quality-assured resource for constructing a diabetes knowledge base. 

\subsection{Annotation Guide}
Two seasoned endocrinologists designed the annotation guide. The guide focuses on entities and relations since these two types are the fundamental elements of a knowledge graph. 

\subsubsection{Entity} 18 types of entities are defined(Table.\ref{tab1}). Nested entities are allowed; for example, \begin{CJK}{UTF8}{gkai}'2型糖尿病'\end{CJK} is a 'Disease' entity, and \begin{CJK}{UTF8}{gkai}'2型'\end{CJK} is a 'Class' one. Entities in DiaKG has two characteristics that stand out: 1. Entities may attribute to different types according to the contextual content; for example, \begin{CJK}{UTF8}{gkai}'糖尿病'\end{CJK} in sentence \begin{CJK}{UTF8}{gkai}'糖尿病患者需控制饮食'\end{CJK} is a 'Disease' type, while in the sentence \begin{CJK}{UTF8}{gkai}'糖尿病所致肾损伤占1/3'\end{CJK} serves as a 'Reason' type; 2. Some entity types are of long spans, like 'Pathogenesis' type is usually consisted of a sentence.

\begin{table}
    \setlength{\abovecaptionskip}{0.cm}
    \setlength{\belowcaptionskip}{-0.cm}
	\caption{List of entities}\label{tab1} 
	\begin{tabular}{|l|l|l|l|}
		\hline
		entity name	& example & \# num & avg length\\
		\hline
		\begin{CJK}{UTF8}{gkai}疾病\end{CJK}(Disease)
		 & \begin{CJK}{UTF8}{gkai}运动对\textbf{\underline{1型糖尿病微血管病变}}的预后无改善作用\end{CJK} & 5,743 & 7.3\\
		\begin{CJK}{UTF8}{gkai}疾病分期分型\end{CJK}(Class)	& \begin{CJK}{UTF8}{gkai}心功能\textbf{\underline{Ⅲ-IV级}}、终末期肾病\end{CJK} & 1,262 & 4.3\\
		\begin{CJK}{UTF8}{gkai}病因\end{CJK}(Reason) & \begin{CJK}{UTF8}{gkai}若\textbf{\underline{体重增加}}，可能加重胰岛素抵抗 \end{CJK} & 175 & 7.3\\
		\begin{CJK}{UTF8}{gkai}发病机制\end{CJK}(Pathogenesis)
		& \begin{CJK}{UTF8}{gkai}多数患者的\textbf{\underline{$\beta$细胞完全破坏}}\end{CJK} & 202 & 10.3\\
		\begin{CJK}{UTF8}{gkai}临床表现\end{CJK}(Symptom)
		& 
		\begin{CJK}{UTF8}{gkai}已发生明确的\textbf{\underline{足趾、足掌坏疽创面}}\end{CJK} & 479 & 5.8\\
		\begin{CJK}{UTF8}{gkai}检查方法\end{CJK}(Test)
		& 
		\begin{CJK}{UTF8}{gkai}进行\textbf{\underline{混合餐耐量试验(MMTT)}}\end{CJK} & 489 & 6.1\\ 
		\begin{CJK}{UTF8}{gkai}检查指标\end{CJK}(Test\_Items)
		& 
		\begin{CJK}{UTF8}{gkai}测量\textbf{\underline{指血(毛细血管血)血糖}}\end{CJK} & 2,718 & 7.7\\
		\begin{CJK}{UTF8}{gkai}检查指标值\end{CJK}(Test\_Value)
		& 
		\begin{CJK}{UTF8}{gkai}血糖\underline{$\bm{<3.3mmol/L}$}\end{CJK} & 1,356 & 9.5\\
		\begin{CJK}{UTF8}{gkai}药物名称\end{CJK}(Drug)
		& 
		\begin{CJK}{UTF8}{gkai}包括\textbf{\underline{COX-2抑制剂}}\end{CJK}  & 4,782 & 7.8\\
		\begin{CJK}{UTF8}{gkai}用药频率\end{CJK}(Frequency)
		& 
		\begin{CJK}{UTF8}{gkai}按照0.5mg，\textbf{\underline{1～3次／d}}\end{CJK} & 156 & 4.7\\
		\begin{CJK}{UTF8}{gkai}用药剂量\end{CJK}(Amount)
		& 
		\begin{CJK}{UTF8}{gkai}可根据\textbf{\underline{0.3～0.5单位/千克体重}}来估算\end{CJK} & 301 & 6.7\\
		\begin{CJK}{UTF8}{gkai}用药方法\end{CJK}(Method)
		& 
		\begin{CJK}{UTF8}{gkai}短效胰岛素一般在\textbf{\underline{餐前15～30min皮下注射}}\end{CJK} & 399 & 6.1\\
		\begin{CJK}{UTF8}{gkai}非药治疗\end{CJK}(Treatment)
		& 
		\begin{CJK}{UTF8}{gkai}\textbf{\underline{认知-行为及心理干预}}是调整患者的生活环境\end{CJK} & 756 & 8.0\\
		\begin{CJK}{UTF8}{gkai}手术\end{CJK}(Operation)
		& 
		\begin{CJK}{UTF8}{gkai}进行\textbf{\underline{胰岛细胞移植手术}}来改善胰岛情况\end{CJK} & 133 & 9.0\\
		\begin{CJK}{UTF8}{gkai}不良反应\end{CJK}(ADE)
		& 
		\begin{CJK}{UTF8}{gkai}贝特类可使\textbf{\underline{胆结石的发生率升高}}\end{CJK} & 874 & 5.1\\
		\begin{CJK}{UTF8}{gkai}部位\end{CJK}(Anatomy) &
		\begin{CJK}{UTF8}{gkai}\textbf{\underline{微血管}}和\textbf{\underline{大血管}}并发症等方面的证据\end{CJK} & 1,876 & 3.1\\
		\begin{CJK}{UTF8}{gkai}程度\end{CJK}(Level)
		& 
		\begin{CJK}{UTF8}{gkai}对于\textbf{\underline{中到重度}}肾功能不全患者需减少剂量\end{CJK} & 280 & 2.9\\
		\begin{CJK}{UTF8}{gkai}持续时间\end{CJK}(Duration)  &
		\begin{CJK}{UTF8}{gkai}预防治疗维持\textbf{\underline{3～6个月}}\end{CJK} & 69 & 3.7\\
		\hline
	\end{tabular}
\end{table}

\subsubsection{Relation}
Relations are centered on 'Disease' and 'Drug' types, where a total of 15 relations are defined(Table.\ref{tab2}). Relations are annotated on the paragraph level, so entities from different sentences may form a relation, which has raised the difficulty for the relation extraction task. Head entity and tail entity existing in the same sentence only account for $43.4\%$ in DiaKG. 

\begin{table}
    \setlength{\abovecaptionskip}{0.cm}
    \setlength{\belowcaptionskip}{-0.cm}
	\caption{List of relations}\label{tab2} 
	\begin{tabular}{|l|l|l|}
		\hline
		relation & example & \# num\\
		\hline
 		TestItems\_Disease  & \begin{CJK}{UTF8}{gkai}血浆\underline{\textbf{酮体}}增加或\underline{\textbf{酮血症}}倾向低于正常人\end{CJK} & 1,171\\
	    Treatment\_Disease    & \begin{CJK}{UTF8}{gkai}积极进行\textbf{\underline{糖尿病}}防治知识的宣教，\textbf{\underline{增加运动}}\end{CJK} & 354\\
 		Class\_Disease         & \begin{CJK}{UTF8}{gkai}分级\textbf{\underline{l-II级}}的\textbf{\underline{充血性心力衰竭}}的患者\end{CJK} & 854\\
		Anatomy\_Disease      &  \begin{CJK}{UTF8}{gkai}慢性开发症如各种\underline{\textbf{\textbf{\underline{神经}}病变}}、视网膜病变等\end{CJK} & 195\\
		Drug\_Disease         & \begin{CJK}{UTF8}{gkai}\textbf{\underline{二甲双胍}}可有效改善\textbf{\underline{糖尿病}}的IR\end{CJK} & 1,315\\
		Reason\_Disease       & \begin{CJK}{UTF8}{gkai}\textbf{\underline{慢性梗阻}}可引起\textbf{\underline{肾积水}}和肾实质萎缩\end{CJK}  & 164\\
		Symptom\_Disease      & \begin{CJK}{UTF8}{gkai}对\textbf{\underline{糖尿病足溃疡}}及...更好地体现了\textbf{\underline{创面感染}}的情况\end{CJK} & 283\\
 		Operation\_Disease     & \begin{CJK}{UTF8}{gkai}接受\textbf{\underline{糖尿病}}外科手术患者...对接受\textbf{\underline{减重代谢手术}}的病人\end{CJK} & 37\\
		Test\_Disease & \begin{CJK}{UTF8}{gkai}5项检查(...\textbf{\underline{温度觉}})等方法半定量评估患者的\textbf{\underline{神经病变}}程度\end{CJK} & 271\\
		Pathogenesis\_Disease & \begin{CJK}{UTF8}{gkai}二甲双胍可改善\textbf{\underline{IR}}...更全面针对\textbf{\underline{T2DM}}的生理缺陷的特点\end{CJK} & 130\\
		ADE\_Drug         & \begin{CJK}{UTF8}{gkai}正确使用\textbf{\underline{磺脲类药物}}...，\textbf{\underline{轻、中度低血糖}}发生率为...\end{CJK} & 693\\
		Amount\_Drug          & \begin{CJK}{UTF8}{gkai} \textbf{\underline{二甲双胍}}(\textbf{\underline{1000mg/d}})起始治疗\end{CJK} & 195\\
		Method\_Drug          & \begin{CJK}{UTF8}{gkai}\textbf{\underline{短效胰岛素}}一般在\textbf{\underline{餐前15～30min皮下注射}}\end{CJK} & 185\\
		Frequency\_Drug       & \begin{CJK}{UTF8}{gkai}\textbf{\underline{每日1次}}基础\textbf{\underline{胰岛素}}或...作为胰岛素起始治疗方案\end{CJK} & 103\\
		Duration\_Drug        & \begin{CJK}{UTF8}{gkai}\textbf{\underline{持续}}静脉泵注\textbf{\underline{胰岛素}}有利于减少血糖波动\end{CJK} & 61\\
		\hline
	\end{tabular}
\end{table}



\subsection{The Annotation Process}
\begin{figure}[]
    \setlength{\abovecaptionskip}{0.cm}
    \setlength{\belowcaptionskip}{-0.cm}
	\centering
	\includegraphics[width=10cm,height=7cm]{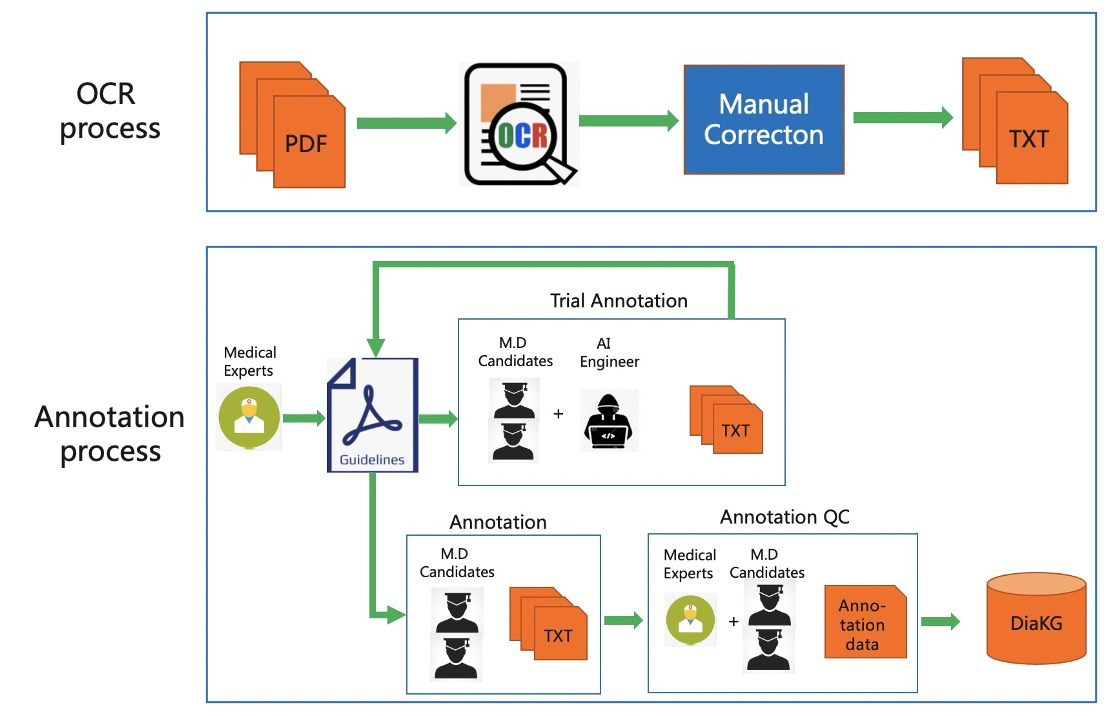}
	\caption{The annotated process of the diabetes dataset.}\label{fig1}
\end{figure}

The annotated process is shown in Fig.\ref{fig1}. The process can be divided into two steps:

\subsubsection{OCR Process}
The PDF files are transformed to plain text format via the OCR tool\footnote[1]{https://duguang.aliyun.com/}, where non-text data like figures and tables are manually removed. Additionally 2 annotators manually check the OCR results character by character to avoid misrecognitions, for example, \begin{CJK}{UTF8}{gkai}'$\beta$细胞'\end{CJK} may be recognized as \begin{CJK}{UTF8}{gkai}'B细胞'\end{CJK}. 

\subsubsection{Annotation Process}
6 M.D. candidates were employed and were trained thoroughly by our medical experts to have a comprehensive understanding of the annotation task. During the \textbf{trial annotation} step, we creatively invited 2 AI experts to label the data simultaneously, based on the assumption that AI experts could provide data insight from the model's perspective. For example, medical experts are inclined to label \begin{CJK}{UTF8}{gkai}'成年型糖尿病(maturity-onset diabetes of the young，MODY)'\end{CJK} as a whole entity, while AI experts regard \begin{CJK}{UTF8}{gkai}'成年型糖尿病'\end{CJK}, 'maturity-onset diabetes of the young' and 'MODY' as three separate entities are more model-friendly. Feedback from AI experts and the annotators were sent back to the medical expert to refine the annotation guideline iteratively. The \textbf{formal annotation} step started by the 6 M.D. candidates and 1 medical experts would give timely help when needed. \textbf{The Quility Control (QC)} step was conducted by the medical experts to guarantee the data quality, and common annotation problems were corrected in a batch mode. The final quality is evaluated by the other medical expert via random sampling of 300 records. The accuracy rates of entity and relation are $90.4\%$ and $96.5\%$, respectively, demonstrating the high-quality of DiaKG. The examined dataset contains 22,050 entities and 6,890 relations, which is empirically adequate for a specified disease. 

\subsection{Data Statistic}
Detailed statistical information for DiaKG is shown in Table.\ref{tab1} and Table.\ref{tab2}.

\section{Experiments}
We conduct Named Entity Recognition(NER) and Relation Extraction(RE) experiments to evaluate DiaKG. The codebase is public on github\footnote[1]{https://github.com/changdejie/diaKG-code}, and the implementation details are also illustrated on the github repository. 

\subsection{Named Entity Recognition (NER)}
We only report results from X Li et al.(2019)\cite{BERT-MRC} since it is the SOTA model for NER with nested settings at the time of this writting.


\subsection{Relation Extraction (RE)}
The RE task is defined as giving the head entity and the tail entity, to classify the relation type. Many sophisticated methods\cite{Zhang} for RE have been proposed recently, due to the simplified setting, we report results from bi-directional GRU-attention\cite{BiGRU-Att} in this paper.   

\begin{minipage}{\textwidth}
	\begin{minipage}[t]{0.45\textwidth}
		\centering
		\makeatletter\def\@captype{table}\makeatother\caption{selected NER results} \label{tab3}
		\begin{tabular}{cccc}        
			\hline
			Entity       & precision & recall & F1 \\
			\hline
			total        & 0.814  & 0.853    & 0.833 \\
			\hline
			Drug         & 0.881   & 0.902   & 0.892 \\
			Disease      & 0.794   & 0.91    & 0.848 \\
			Pathogenesis & 0.595   & 0.667   & 0.629 \\
 			Symptom      & 0.535   & 0.535   & 0.535 \\
            Reason       & 0.333   & 0.3     & 0.316 \\
		\end{tabular}
	\end{minipage}
	\begin{minipage}[t]{0.45\textwidth}
		\centering
		\makeatletter\def\@captype{table}\makeatother\caption{selected RE results} \label{tab4}
		\begin{tabular}{cccc} 
			\hline
			Relation & precision & recall & F1 \\
			\hline 
			total &	0.839 & 0.837 & 0.836 \\
			\hline
			Class\_Disease        & 0.968 & 0.874 & 0.918 \\
			ADE\_Drug             & 0.892 & 0.892 & 0.892 \\
			Test\_Disease         & 0.648 & 0.636 & 0.642 \\			
			Pathogenesis\_Disease & 0.486 & 0.692 & 0.571 \\
			Operation\_Disease     & 0.6   & 0.231 & 0.333 \\
		\end{tabular}
	\end{minipage}
\end{minipage}

\section{Analysis}

The experimental results are shown in Table.\ref{tab3} and Table.\ref{tab4}. We report the total result, plus the top 2 and last 3 types' results for each task to analyze DiaKG.

The \textbf{overall} macro-average scores for the two tasks are $83.3\%$ and $83.6\%$, respectively, which are satisfying considering the multifarious types we define, also demonstrating DiaKG's high quality. For the \textbf{NER task}, the results of 'Disease' and 'Drug' types are as expected because these two types exist frequently among the documents, thus leading to a higher score. The average entity length for 'Pathogenesis' type is 10.3, showing that the SOTA MRC-Bert model still can not handle the long spans perfectly; We analyzed errors of the 'Symptom' and 'Reason' types and found that the model is prone to classify entities as other types, mainly contributing to the characteristic that entity may be of different types due to the contextual content. For the \textbf{RE task}, the case study shows that entities with long distance are difficult to classify. For example, entities with 'Drug\_Diesease' type usually exist in the same sub-sentence, whereas the ones with 'Reason\_Disease' type are usualy located in different sub-sentences, sometimes even in different sentences. The above experimental results demonstrate that DiaKG is challenging for most current models and it is encouraged to employ more powerful models on this dataset.



\section{Conclusion $\&$ Future Work}
In this paper, we introduce DiaKG, a specified dataset dedicated to the diabetes disease. Through a carefully designed annotation process, we have obtained a high-quality dataset. The experiment results prove the practicability of DiaKG as well as the challenges for the most recent typical methods. We hope the release of this dataset can advance the construction of diabetes knowledge graphs and facilitate AI-based applications. We will further explore the potentials of DiaKG and contribute it to the CBLUE\cite{CBLUE} community. 

\section{Acknowledgments}
We want to express gratitude to the anonymous reviewers for their hard work and kind comments. We also thank Tianchi Platform to host DiaKG.
%
%

%
%
%

\begin{thebibliography}{88}

\bibitem{BMJ}
Li Y ,  Teng D ,  Shi X , et al. Prevalence of diabetes recorded in mainland China using 2018 diagnostic criteria from the American Diabetes Association: national cross sectional study[J]. BMJ, 2020, 369.

\bibitem{J}
Luo Z , Fabre G , Rodwin V G . Meeting the Challenge of Diabetes in China[J]. International Journal of Health Policy and Management, 2020, 9(2).

\bibitem{CDIA}
Holmes, David. Linong Ji: fighting to turn the tide against diabetes in China[J]. Lancet, 2014, 383(9933):1961-1961.

\bibitem{KG}
Nickel, M. et al. A Review of Relational Machine Learning for Knowledge Graphs. Proceedings of the IEEE 104.1(2015):11-33

\bibitem{KNEE}
Bisson LJ, Komm JT, Bernas GA, et al. Accuracy of a computer-based diagnostic program for ambulatory patients with knee pain. The American journal of sports medicine 2014;42(10):2371-6.

\bibitem{KGE}
WangM, LiuM, LiuJ, et al. Safe medicine recommendation via medical knowledge graph embedding. arXiv preprint arXiv:1710.05980.2017.

\bibitem{TANGH}
Tang, H. , and  J. Ng . "Googling for a diagnosis\begin{CJK}{UTF8}{gkai}—\end{CJK}use of Google as a diagnostic aid: internet based study." BMJ 333(2006).

\bibitem{GANN}
Gann B. Giving patients choice and control: health informatics on the patient journey.[J]. Yearb Med Inform, 2012, 21(01):70-73.

\bibitem{Zhang}
N. Zhang et al., "Contrastive Information Extraction with GenerativeTransformer," in IEEE/ACM Transactions on Audio, Speech, and Language Processing, doi: 10.1109/TASLP.2021.3110126.


\bibitem{BERT-MRC}
X  Li,  Feng J ,  Meng Y , et al. A Unified MRC Framework for Named Entity Recognition[J].  2019.

\bibitem{BiGRU-Att}
Peng Z ,  Wei S ,  Tian J , et al. Attention-Based Bidirectional Long Short-Term Memory Networks for Relation Classification[C]// Proceedings of the 54th Annual Meeting of the Association for Computational Linguistics (Volume 2: Short Papers). 2016.

\bibitem{CBLUE}
Zhang, N., Bi, Z., Liang, X., Li, L., Chen, X., Deng, S., Li, L., Xie, X., Ye, H., Shang, X., Yin, K., Tan, C., Xu, J., Chen, M., Huang, F., Si, L., Ni, Y., Xie, G., Sui, Z., Chang, B., Zong, H., Yuan, Z., Li, L., Yan, J., Zan, H., Zhang, K., Chen, H., Tang, B., and Chen, Q. (2021). CBLUE: A Chinese Biomedical Language Understanding Evaluation Benchmark. ArXiv, abs/2106.08087.

\end{thebibliography}
%

\end{document}